\documentclass{elsarticle}

\usepackage{graphicx}
\usepackage[colorlinks]{hyperref}
\usepackage{amsthm,amssymb,amsmath}
\usepackage{comment}
\usepackage{setspace} 
\theoremstyle{plain}
\newtheorem{theorem}{Theorem}[section]
\newtheorem{lemma}[theorem]{Lemma}
\newtheorem{proposition}[theorem]{Proposition}
\newtheorem{corollary}[theorem]{Corollary}

\theoremstyle{definition}
\newtheorem{definition}{Definition}[section]

\title{An Upper Bound for Minimum True Matches in Graph Isomorphism with Simulated Annealing}

\author{Hashem Ezzati$^1$  and   Mahmood Amintoosi$^{2,*}$ and Hashem Tabasi$^{3}$ \\
$^1$Amirkabir University of Technology, Tehran, IRAN, h.ezzati@aut.ac.ir\\
$^2$Hakim Sabzevari University, Sabzevar, IRAN, m.amintoosi@hsu.ac.ir\\
$^3$Damghan University, Sabzevar, IRAN}

\begin{document}

\begin{abstract}
Graph matching is one of the most important problems in graph theory and combinatorial optimization, with many applications in various domains. Although meta-heuristic algorithms have had good performance on many NP-Hard and NP-Complete problems, for this problem there are not reported superior solutions by these algorithms. The reason of this inefficiency  has not been investigated yet. In this paper we  show that simulated annealing as an stochastic optimization method is unlikely to be even close to the optimal solution for this problem. In addition to theoretical discussion, the experimental results also verified our idea; for example, in two sample graphs, the probability of reaching to a solution with more than three correct matches is about $0.02$ in simulated annealing.\\
\textbf{keywords: Graph matching, Simulated Annealing, Meta-Heuristic, Stochastic Optimization}
\end{abstract}

\maketitle
%

\section{Introduction}
Several important applications in computer vision, such as 
3D reconstruction from stereo images , object matching,
object category and action recognition 
\cite{Szeliski:2010:CVA} 
require the ability to efficiently match features of two sets. 
Image features are considered as graph vertices for modeling the matching problem as a graph matching problem.

\begin{figure}
	\centering
	\includegraphics[width=1\linewidth]{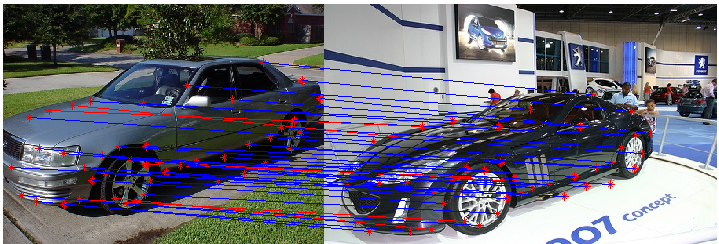}
	\caption{Two car images consider as graphs and thier Correspondence that incorrect matches are shown in red}
	\label{fig:imageMatching1}
\end{figure}

Graph matching is one of the oldest problems in Graph theory and combinatorial optimization, 
which has attracted so much attention to itself. It also
 has various applications such as 
Image matching in computer vision
\cite{Kapochino2010,pishraft2016}. Consider two graphs with $n$ vertices in each graph, the number of all possible matching cases between them is $n!$. There are several variants of the graph matching problems, one of them is matching in Isomorphic Graphs. An example of this variant is 2D Image Matching Problem. Figure \ref{fig:imageMatching1} shows a pair of car images along with their corresponding graphs. The blue/red lines represent true/false matches.

Graph matching is an NP-hard problem \cite{pishraft2016} with complexity of $O(n!)$ in its general form. A permutation of numbers $\{1, \dots, n\}$ shows a point in the solution space.
In many applications such as figure 
\ref{fig:imageMatching1} only
just one permutation  among $ n!$ possible solutions is acceptable. Meta-heuristic algorithms which are mainly based on  random search are used on many NP-hard and NP-complete problems successfully. For example, they are efficient on Traveling Salesman, N-Queen, Facility Location, Graph Coloring, Max-Cut Problem, bin-packaging, time tabling and etc. \cite{hendi,Tohyama2011GA,nqueenGA}. 



There are several methods including linear programming, spectral, eigen decomposition, genetic algorithm and simulated annealing to solve the graph matching problem. We divided the published related works to journal and non-journal papers:

\begin{description}
	\item[Journal Papers:] 
Egozi et. al. \cite{Egozi2013} reported an spectral approach for graph matching. 	
Qiao et. al. \cite{Spectral5} also used spectral graph matching in remote sensing.
Finch et.al. \cite{Em2} and Luo and Hancock \cite{Em1} reported Expectation Maximization and Singular Value Decomposition methods for dealing the problem.
Almohamad and Duffuaa \cite{1993Al} used linear programming.
Gradient descent approach was used in \cite{Williams1999Deterministic}.
Shapiro and Brady \cite{Spectral3} and Umeyama \cite{Spectral1} solve the matching problem by eigen decomposition.
Brown et.al. \cite{BrownJWG94}, 
Auwatanamongkol \cite{AUWATANAMONGKOL20071428}, 
Cross et. al. \cite{CROSS1997953,CROSS20001863} and Fr\"{o}hlich and Ko\v{s}ir \cite{FROHLICH2001195} used genetic algorithm (GA).

	\item[Non-Journal Papers:] 
	
Leordeanu and Hebert \cite{Spectral2005} reported spectral method.
Using simulated annealing (SA) \cite{Sima},
ant colony optimization \cite{Ant2005}
and tabu search (TS) \cite{Kapochino2010} also reported by the researchers. 
In 
 \cite{Choi:2012:EGA,Wang1998GA} also GA has been used for the problem.
\end{description}

As can be seen, meta-heuristic approaches -- except GA -- were only reported in non-journal published papers.
In this paper we will discuss why this sort of approaches are not efficient to solve matching problem.

 Some meta-heuristic approaches like SA and TS are based on defining a neighboring structure and move from one point in solution space to another point.
Previous journal papers shows that stochastic optimization methods like SA are not attractiveness for researchers. 
This fact and the authors failure to solve the matching problem by SA, makes this intuitive result that methods like as SA are not effective for matching problem.
GA uses crossover in addition to mutation, here the performance of this algorithm is not investigated.

In this study, we will show that the probability of reaching the optimal solution (success rate) is close to zero, with simulated annealing. Simulated annealing is a kind of an stochastic optimization methods;  these methods start from a random point in the solution space and produce neighbors by  applying random operators.

Section \ref{sec:theo} devoted to theoretical aspects of an upper bound on success rate in SA. Section \ref{sec4}, shows  experimental results on various graphs for inspecting the theoretical results. Conclusion remarks is our last section.

\section{Theoretical Discussion}
\label{sec:theo}

In some problems , there are many optimal points in the solution space,  which makes easier finding the solution. Finding the optimal point, takes more time, when the optimal points in solution space are rare. 
N-Queen problem \cite{dancinglinks} is a problem with so many correct solutions. For example there are about 100 milion correct answers for $n=17$. As can be seen in table \ref{tab:NQueen}, the number of solutions rapidly grows, when $n$ increases.
But in some problems such as graph matching, only one correct solution exist.
Our goal is to prove that the ratio of partly good solutions decreases, when the number of graph vertices increases, for graph matching.
Before bringing into mathematical derivation, some basic definitions are explained here.


\begin{table}[t]
	\caption{Number of solutions of N-Queen Problem \cite{dancinglinks}.}
	\label{tab:NQueen}
	\centering
	\includegraphics[width=.2\linewidth]{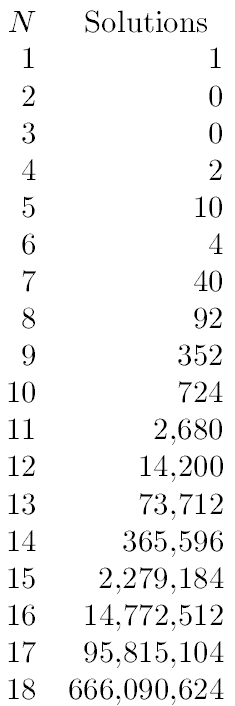}
\end{table}


\subsection{Basic Definitions}
\begin{definition}[Graph Matching]
	Consider two graphs $G_1$ and $G_2$, a one-to-one mapping from$ G_1 $ to $G_2$ represents a matching between the two graphs
	\cite{hendi}.
\end{definition}

\begin{definition}[Isomorphism]
	Consider two graphs $G_1 $ and $ G_2$, they are isomorphic if and only if there is a mapping $\phi:V(G_1)\longrightarrow V(G_2)$ as each vertex of $G_1$ exactly corresponds to a vertex of $G_2$ (Any two vertices x and y of $G_1$ are adjacent in $G_1$ if and only if $\phi(x)$ and $\phi(y)$ are adjacent in $G_2$)\cite{hendi}.
\end{definition}

\begin{description}
	\item[Simulated Annealing Algorithm:]
SA algorithm starts from a random point, and picks a random move to go to a new point in the neighbor of the previous point.  If the new point improves the solution (decreases energy), then it is always accepted.  Otherwise, it is accepted with some probability.  This probability decreases exponentially over time and with the badness of the move.

\item[Feasible Solution:]  A feasible solution or a point in the solution space is a permutation of $\{1,\dots,n\}$, in matching problem.

\item [True Solution:] In this paper it is supposed that the true solution is numbers $\{1,\dots,n\}$ in ascending  order.
\item[Energy (Error):] 
The number of elements that are not in their true positions, is the solution energy (error), which should be minimized. This measure is normalized by dividing by $n$.
For example suppose that the true solution of a matching problem be \{1,2,3,4,5\} and a random solution be \{3,2,1,4,5\}. 
In this permutation  $1, 3$ are not in their right places, 
hence its corresponding error is $2/5$.
In stochastic optimization methods like SA, in each run, the starting point is selected randomly; the mean error is computed by averaging the errors over multiple runs. 
	\item [Precision:] Precision is 1 minus error.
	\item[Derangement:] a  derangement is a permutation of the elements of a set, such that no element appears in its original position. The number of derangements of a set of size n, usually written $D_n$, $d_n$, or $!n$, is called the ``derangement number".
\end{description}

In the following section we will show that precision, approaches zero, when the graph size increases.

 
\subsection{Theorems}
Each feasible solution is a random permutation of \{1, \dots, n\}, in stochastic optimization approaches like SA, the new solution is produced by perturbing the current solution. The common methods for permutation perturbation are
 swapping, insertion, inversion and scramble operators. All of these operators make a new vector (solution) from the current vector by altering some part of the vector. In fact, moving to the new point is just moving to a new random point, around the previous point. Here we discuss the probability of ``goodness" of a random point.

In a solution space of permutations vectors, there are $n!$ points.
Remmel \cite{Remmel83} showed that the number of permutations of $\{1,\dots,n\}$, that no number is located in its true position (known as derangement) is:

\begin{align}
D(n) &= (n-1)(D(n-1) + D(n-2)), \\
&  n\ge 1, D(0)=1, D(1)=0 \nonumber
\label{for1.1}
\end{align}
which is equivalence to following non-recursive summation \cite{Remmel83}:
\begin{equation}
D(n) =  n! \sum_{k=0}^{n} \frac{-1^k}{k!}
\label{for1.2}
\end{equation}

For demonstrating that the SA move operator is week to produce good results, it is sufficient to show that the probability of producing good solutions is very low with this manner. In the following it is shown that the expected value for having a solution with only 3 correct matches  is about 0.02. Before that, some preliminary theorems should be explained.

\begin{lemma}
The number of permutations of $\{1,\dots,n\}$, that exactly $m<n$ elements are located in their true positions is:
$$G(n,m)=\left(\begin{array}{c}n\\ m\end{array}\right)D(n-m)$$
\label{theorem2}
\end{lemma}
 
 \begin{proof}
The proof is obvious, if $m$ elements are located correctly, $n-m$ elements are not on their true positions. derangement number of these elements is $D(n-m)$. Since the number of $m$-combinations of $n$ elements is $\left(\begin{array}{c}n\\ m\end{array}\right)$, the total number of the mentioned permutations is:
 	
   $$G(n,m)=\left(\begin{array}{c}n\\ m\end{array}\right)D(n-m)$$
  \end{proof}
 

In the following, the probability of being $m<n$ elements in their true positions, denoted by $P^n_m$.

\begin{theorem}
The probability of being $m<n$ elements of $n$ elements on their true positions in a random permutation -- denoted by  $P^n_{m}$ -- is $\frac{ D(n-m)}{ m! \times (n-m)!}$.
\label{theorem4}
\end{theorem}

\begin{proof}
The total points in the solution space is $n!$; by using Lemma \ref{theorem2} we have: 
 \begin{align}
 P^n_m&= \frac{G(n,m)}{ n!}\nonumber\\
 &=\frac{\left(\begin{array}{c}n\\ m\end{array}\right)D(n-m)}{n!}
 =\frac{n! \times D(n-m)}{n!\times m! \times (n-m)!}\nonumber\\
 &=\frac{ D(n-m)}{ m! \times (n-m)!}\nonumber
 \end{align}
\end{proof}

\begin{corollary}
$\lim_{n\to\infty} P^n_m =\frac{e^{-1}}{m!}$.
\label{corollary1.1}
\end{corollary}
\begin{proof}
 \begin{align}
\lim_{n\to\infty} P^n_m&=\lim_{n\to\infty} \frac{ D(n-m)}{ m! \times (n-m)!}\nonumber\\
 &=\frac{1}{ m!}\times \lim_{n\to\infty} \frac{ D(n-m)}{(n-m)!}\nonumber
 \end{align}

Substituting $n-m$ by $L$ and using eq. \eqref{for1.2} yields:
 \begin{align}
\lim_{n\to\infty} P^n_m &=  \frac{1}{ m!}\times \lim_{L\to\infty} \frac{ L!}{L!} \sum_{k=0}^{L} \frac{-1^k}{k!}\nonumber
\end{align}
since:
$$e^x = \sum_{n = 0}^{\infty} {x^n \over n!}$$
hence:
 \begin{align}
\lim_{n\to\infty} P^n_m &=  \frac{e^{-1}}{m!}
\label{for222}
\end{align}
\end{proof}

%
%

\begin{corollary}
$\lim_{n\to\infty}P^n_0$ is ${e}^{-1}$.
\label{corollary1.2}
\end{corollary}

\begin{corollary}
	For large $n$, the probability of being \textbf{at most} $m$ elements of $n$ elements in their true positions
	 is:
\[e^{-1}+\frac{e^{-1}}{1!}+\frac{e^{-1}}{2!}+...+\frac{e^{-1}}{m!}\]	
\end{corollary}	

\begin{corollary}
	For large $n$, the probability of being \textbf{at least} $m$ elements of $n$ elements in their true positions
	-- denoted by $P^n_{1:m}$ -- 
	is:
	\begin{equation}
	P^n_{1:m}=1-(e^{-1}+\frac{e^{-1}}{1!}+\frac{e^{-1}}{2!}+...+\frac{e^{-1}}{m!})
	\label{formolfarei5}
	\end{equation}
\end{corollary}	

\begin{proposition}
From the last corollary it is obvious that the probability of being $3$ of $n$ elements on their correct positions 
is $0.02$:

\begin{align}
		P^n_{1:3}&=1-(e^{-1}+\frac{e^{-1}}{1!}+\frac{e^{-1}}{2!}+\frac{e^{-1}}{3!})\nonumber\\
		&=1-(e^{-1}\times(1+\frac{1}{1!}+\frac{1}{2!}+\frac{1}{3!}))\nonumber\\
		&=1-(e^{-1}\times(2.6666))\approx0.02 \nonumber
\end{align}		
	\label{prop:P_1:3}

\end{proposition}

Suppose that the goodness of a solution (its precision) is the total number of its elements ($m$) that are located on their true positions . 
The above inference showed that, as $m$ increases, the probability to have a partly good solution decreases.
In other words the probability to have at least $m$ true matches in a solution vector is bounded by a small number, which makes SA non-effective to solve graph matching problem.

\begin{table}[t]
	\centering
	\scalebox{0.9}
	{
		\begin{tabular}{|p{18mm}|p{35mm}|p{12mm}|}
			\hline
			Number of Vertices ($n$) &Number of permutations with at least 3 elements in their true positions & $P^n_{1:3}$
			\\
			\hline
			$20$  &   $1827$  &   $0.0183$  \\ 
			$50$    &   $1856$  &   $0.0186$  \\ 
			$100$  &   $1874$  &   $0.0187$  \\ 
			$300$  &   $1943$  &   $0.0194$ \\ 
			$500$  &   $1872$  &   $0.0187$ \\
			$1000$  &   $1939$  &   $0.0193$ \\
			$10000$  &   $1862$  &   $0.0186$ \\
			\hline
	\end{tabular}}
	\caption{
		The probability of being at least 3 elements on their true positions for different values of $n$. In each row 100000 permutations were generated randomly.
	}
	
	\label{table1}
	
\end{table}

\section{Experimental Results}\label{sec4}
The previous theoretical discussion is verified by experimental results in this section.
The experiments performed in this section are divided into two parts.
In the first part, a large number of random permutations is generated and the correctness of proposition \ref{prop:P_1:3} is shown.
In second part, this is verified by applying SA algorithm on random graphs.


\subsection{Random permutations results}
Here 100000 random permutations for various values of $n$ are generated. For example for $n=100$, 100000 random permutation is generated and the number of those permutations that at least 3 elements are located on their true positions is counted. The results is shown in table \ref{table1}. The last column of the table shows $P^n_{1:3}$, which according to proposition \ref{prop:P_1:3} should be equal 0.02. As can be seen the experimental results verifies theory.


\vspace{-5mm}
\subsection{Simulated Annealing Results}
Simulated annealing runs on the random graph analogous to the previous section. Table \ref{table2}  shows the simulation results. Each $n$ corresponds to vertices numbers of two graphs with equal number of nodes, which should be matched.
The graphs are produced according to \cite{Kapochino2010}.
 Theoretical deductions is satisfied for those graphs having large number of nodes. 
 As can be seen, as the number of vertices increases, the value of the last column approaches to 0.02, which verifies proposition \ref{prop:P_1:3}.



\begin{table}[t]
	\centering
	\scalebox{0.9}
{
\begin{tabular}{|p{3cm}|p{15mm}|}
			\hline
Number of vertices & $P^n_{1:3}$
\\
			\hline
        $20$    &   $0.40$   \\
		$50$    &    $0.16$  \\
		$100$ &    $0.02$  \\
		$300$   &    $0.04$   \\
		$500$   &    $0.024$  \\
		$1000$  &    $0.019$  \\
		$10000$ &    $0.02$ \\
			\hline
		\end{tabular}}
		\caption{As the number of vertices increases, the value of the second column approaches to 0.02, which verifies propos. \ref{prop:P_1:3}.}
\label{table2}
\end{table}

\section{Conclusions}

In this study, an upper bound for minimum true matches in graph matching problem using simulated annealing has been proposed. 
Graph matching problem for some problems like image stereo matching, has just one optimal solution. We argued that finding the optimal solution with stochastic optimization methods are not feasible, due to the low probability of the number of true matches.
Mathematical derivation showed that the rate of true matches is very low, in a random based search method. For example the probability that only 3 nodes were matched correctly is about 0.02. Experimental results verified theoretical discussion.

\vspace{5mm} 

\noindent\textbf{Compliance with ethical standards}\\
\noindent\textbf{Conflict of interest} The authors declare that they have no conflict of interest.\\
\noindent\textbf{Ethical approval:} This article does not contain any studies with human participants or animals performed by any of the authors.


\bibliographystyle{spmpsci}      
\bibliography{MyReferences}   

\end{document}